\documentclass[conference]{IEEEtran}
\ifCLASSINFOpdf
  \usepackage[pdftex]{graphicx}
\else
\fi

\usepackage{amsmath}

\usepackage[tight,footnotesize]{subfigure}

\usepackage{fixltx2e}
\usepackage{subfig}
\usepackage{graphicx}
\begin{document}
%
\title{Multiple-Instance Logistic Regression \\with LASSO Penalty}



%
\author{\IEEEauthorblockN{Ray-Bing Chen\IEEEauthorrefmark{1},
    Kuang-Hung Cheng\IEEEauthorrefmark{2},
    Sheng-Mao Chang\IEEEauthorrefmark{1},
    Shuen-Lin Jeng\IEEEauthorrefmark{1},
    Ping-Yang Chen\IEEEauthorrefmark{1}, \\
    Chun-Hao Yang\IEEEauthorrefmark{1}, and
    Chi-Chun Hsia\IEEEauthorrefmark{2}}
    \IEEEauthorblockA{\IEEEauthorrefmark{1}
        Department of Statistics \\ National Cheng Kung University \\
        No. 1, University Road, Tainan, Taiwan}
    \IEEEauthorblockA{\IEEEauthorrefmark{2}
        Industrial Technology Research Institute \\
        ITRI Southern Region Campus\\
         Tainan, Taiwan}
}



\maketitle

\begin{abstract}
In this work, we consider a manufactory process which can be
described by a multiple-instance logistic regression model. In
order to compute the maximum likelihood estimation of the unknown
coefficient, an expectation-maximization algorithm is proposed, and
the proposed modeling approach can be extended to identify the
important covariates by adding the coefficient penalty term into the
likelihood function. In addition to essential technical details, we
demonstrate the usefulness of the proposed method by simulations and
real examples.
\end{abstract}


%
\IEEEpeerreviewmaketitle

\section{Introduction}

We consider the data generated from a stable manufacturing process.
A total of $n$ subjects are obtained, and each subject consists of a
number of components. Along with each component, $p$ predictors are
observed. The anticipated response is the status of the component,
defective or not. However, it is impractical to check the status of
all components within each subject. The status of the subject,
instead, is observed. For a particular subject, if its one or more
components are defective, the subject is defective,  and otherwise
the subject is not defective. The goal of this work is to predict
whether a subject is defective and to identify covariates that
plausibly affect the defect rate especially when the pool of
covariates is very large and only a few of them truly affects the
defect rate.

For the purpose of defect prediction, multiple-instance (MI)
learning \cite{D97} is a solution. The difference between the
traditional supervised learning  and the MI learning is as follows.
In the traditional supervised learning setting, the labels of each
instance (components) are given, while in a typical MI setting,
instances are grouped into bags (subjects) and only the labels of
each bag are known, i.e. labels for the instance are missing. That
is, we do not have the complete data for model fitting. To analyze
MI data, the relationship between the instances and bags must be
explicitly posited. Most of the research on MI learning is based on
the standard MI assumption \cite{F10} which assumes that a positive
bag contains at least one positive instance while a negative bag
contains no positive instances and all instances in a negative bag
must be negative. This assumption is hold throughout this article.
Many methods have been proposed for MI learning. Most of these
methods are extensions of support vector machine and logistic
regression. Other methods such as  Diverse Density \cite{M98} and
EM-DD \cite{ZG01} are also feasible.

The first goal of this study focuses on using logistic regression to
model MI data. This method is named multiple-instance logistic
regression (MILR) in \cite{XF04} and \cite{RC05}. We first fix
notation. Consider an experiment with $n$ subjects (bags). Suppose
that, for the $i$th subject, $m_{i}$ independent components
(instances) are obtained. For the $j$th component of the $i$th
subject, the data consists of binary response $y_{ij}$ and the
corresponding covariates $x_{ij}$, a $p$-dimensional vector. We
model the response-predictor relationship by logistic regression;
that is $Y_{ij}\sim Bernoulli(p_{ij})$ where $p_{ij}=p(\beta_0 +
x_{ij}^{T}\beta)$ with $p(x)=1/(1+e^{-x})$,  $\beta_0$ is a
constant term and $\beta$ is a $p \times 1$ unknown coefficient
vector. However, in this experiment, the labels of instances,
$y_{ij}$'s, are not observable. Instead, the labels of the bags,
$Z_{i}=I(\sum_{j=1}^{m}Y_{ij}>0)$'s, is observed. The logistic
regression for bags is therefore
\begin{equation}\label{eq:Lik}
    Z_{i}\sim Ber(\pi_{i})\quad\mbox{where}\quad\pi_{i}=1-\prod_{j=1}^{m_{i}}( 1-p_{ij})
\end{equation}
with likelihood
$L(\beta_0,\beta)=\prod_{i=1}^{n}\pi_{i}^{z_{i}}(1-\pi_{i})^{1-z_{i}}$.
Directly maximizing $L$ with respect to $\beta$ can be initial-value
sensitive or unstable while the number of missing variables  (the
number of components per subject) increases.

In literature, instead of maximizing the likelihood function
$L(\beta_0,\beta)$ directly, alternative likelihood functions were
applied. Especially,  several functions of $(\beta_0, \beta)$ were
proposed to model $\pi_i$. For example, arithmetic mean and
geometric mean of $\{p_{i1},\dots, p_{im_i}\}$ were used to model
$\pi_i$ in \cite{XF04} whereas the softmax function
\[
    S_i(\alpha)= \sum_{j=1}^{m_i} p_{ij} \exp\left\{\alpha p_{ij}\right\}/ \sum_{j=1}^{m_i} \exp\left\{\alpha p_{ij}\right\}
\]
were used to model $\pi_i$ in \cite{RC05} where $\alpha$ is a
pre-specified nonnegative value. According to the relationship
between the bag and the associated instances,  the geometric, the
arithmetic and the softmax function have the following relationship
\[\begin{split}
    &\exp\left\{ \sum_{j=1}^{m_i} \log(p_{ij}) / n \right\}
    \leq \sum_{j=1}^{m_i} p_{ij}/{m_i}=S_i(0) \\
    &\leq S_i(\alpha) \leq\max_{j=1,\dots, m_i} p_{ij},
    \leq P\left( \cup_{j=1}^{m_i} \left[Y_{ij}=1\right] \right) = \pi_i
\end{split}\]
for all $\alpha > 0$. Consequently, when using the same data, the
resulting maximum likelihood estimates for these link functions
should be different although the estimates of $\pi_i$'s may be
similar.  We conclude that directly tackling the likelihood function
(\ref{eq:Lik}), if possible, is  more relevant than others when
parameter estimation is also an important goal of the experiment.
In order to obtain the maximum likelihood estimates, an
expectation maximization algorithm \cite{D77} is proposed because we
treat the labels of the components as missing variables.

Another goal of this work is to identify important covariates
affecting the defect rate in both the instance and the bag levels
and to predict the rate change when a covariate is changed. This
goal supports the use of (\ref{eq:Lik}) because the regression
coefficient estimate is essential to predict the rate change. When
the number of covariates is large  using the traditional variable
selection tool such as Wald test is not efficient. Alternatively,
maximum likelihood approach with LASSO penalty (Tibshironi, 1996) is
promising. In this work, we incorporate the LASSO approach to the
proposed MILR and provide an efficient computer algorithm for
variable selection and estimation. Finally the important variables
are identified if the corresponding coefficient estimations are
nonzero.

The rest of this article is as follows. In Section 2, we introduce expectation-maximization (EM; \cite{D77}) algorithm to find the maximum likelihood estimator of MILR. In Section 3, we discuss the technical details about how to integrate the  LASSO approach to the MILR. In Section 4, we use simulation to demonstrate the benefit of using MILR in the standpoint of variable selection and parameter estimation in contrast to the naive method and other MILR methods. Finally, in Section 5, we use various datasets to evaluate the proposed method.



\section{Multiple-Instance Logistic Regression with EM Algorithm}
Here, we follow the notation defined in previous section. When the
labels of the instance level, $y_{ij}$'s, are observed, the complete
data likelihood function is
\[
    \prod_{i=1}^n \prod_{j=1}^{m_i} p_{ij}^{y_{ij}}(q_{ij})^{1-y_{ij}}
\]
where $q_{ij}=1-p_{ij}$. However, in MI experiments, $y_{ij}$'s are
not observable and instead the labels of the bag level,
$Z_{i}=I(\sum_{j=1}^{m_i}Y_{ij}>0)$'s, are observed. Under this
circumstance, the naive approach uses the likelihood
\[
    L_{N}(\beta_0, \beta)=\prod_{i=1}^{n}\prod_{j=1}^{m_{i}}p_{ij}^{z_{i}}q_{ij}^{1-z_{i}}.
\]
by setting $y_{ij} = z_i$ for all $j$. The resulting testing
and estimation for $\beta_0$ and $\beta$ is questionable since the
probability model does not fit the underlying data generating
process. The idea of the naive approach is that since the instance
labels are missing, the bag label is used to guess the instance
labels. A better approach to treat missing data is the EM algorithm.

To deliver the E-step, the complete data likelihood and the
conditional distribution of missing data conditional on observed
data are required. The complete data log-likelihood is
straightforward,
\[
    l_{C}(\beta_{0},\beta)=\sum_{i=1}^{n}\sum_{j=1}^{m_{i}}y_{ij}\log(p_{ij})+(1-y_{ij})\log(q_{ij}).
\]
The conditional distribution is discussed under two conditions.
First, when $Z_{i}=0$,
\[
    \Pr(Y_{i1}=0,\dots,Y_{im_{i}}=0|Z_{i}=0)=1,
\]
and second, when $Z_{i}=1$,
\[\begin{split}
    &\Pr(Y_{ij}=y_{ij} \mbox{ for all } j|Z_{i}=1)\\
    &=\frac{\prod_{j=1}^{m_{i}}p_{ij}^{y_{ij}}q_{ij}^{1-y_{ij}}\times I(\sum_{j=1}^{m_{i}}y_{ij}>0)}
    {1-\prod_{l=1}^{m_{i}}q_{il}}.
\end{split}\]
Thus, the required conditional expectations are
\[\begin{split}
& E(Y_{ij}|Z_{i}=0)=0\quad\mbox{and}\\
& E(Y_{ij}|Z_{i}=1)=\frac{p_{ij}}{1-\prod_{l=1}^{m_{i}}q_{il}}\equiv\gamma_{ij}.
\end{split}\]
Because $\gamma_{ij}$ is a function of
$p_{ij}=p(\beta_{0}+x_{ij}^{T}\beta)$, denote
$\gamma_{ij}=\gamma_{ij}(\beta_{0},\beta)$. Consequently, for the
$i$th subject, the $Q$ function in the E-step is
\[\begin{split}
    &Q_{i}(\beta_{0},\beta\;|\;\beta_{0}^{t},\beta^{t}) \\
    & =E\left(\left.\sum_{j}y_{ij}\log(p_{ij})+(1-y_{ij})\log(q_{ij})\right|Z_{i}=z_{i},\beta_{0}^{t},\beta^{t}\right)\\
    & =\left[\sum_{j}\log(q_{ij})\right]^{1-z_{i}}\left[\sum_{j}\gamma_{ij}^{t}\log(p_{ij})+(1-\gamma_{ij}^{t})\log(q_{ij})\right]^{z_{i}}\\
    & =\sum_{j}z_{i}\gamma_{ij}^{t}\log(p_{ij})+(1-z_{i}\gamma_{ij}^{t})\log(q_{ij})\\
    & =\sum_{j}z_{i}\gamma_{ij}^{t}(\beta_0+x_{ij}^{T}\beta)-\log(1+e^{\beta_0+x_{ij}^{T}\beta})
\end{split}\]
where $\gamma_{ij}^{t}=\gamma_{ij}(\beta_{0}^{t},\beta^{t})$ and $\beta_{0}^{t},\beta^{t}$ are the estimate obtained in step $t$. Let $Q(\beta_{0},\beta\;|\;\beta_{0}^{t},\beta^{t})=\sum_{i=1}^{n}Q_{i}(\beta_{0},\beta\;|\;\beta_{0}^{t},\beta^{t})$.

Next, we  move to the M-step, i.e. maximize $Q$ with respect to $(\beta_0,\beta)$.
However, this $Q$ is a nonlinear function of $(\beta_0,\beta)$ and, consequently, the maximization is
computational expensive. Following \cite{F10}, we applied the quadratic approximation to the $Q$ function. By taking Taylor expansion about $\beta_{0}^{t}$ and $\beta^{t}$, we have
\[\begin{aligned}
    & Q(\beta_{0},\beta\;|\;\beta_{0}^{t},\beta^{t}) \\
    =&\sum_{i=1}^{n}\sum_{j=1}^{m_{i}}z_{i}\gamma_{ij}^{t}(\beta_{0}+x_{ij}^{T}\beta)-\log(1+e^{\beta_{0}+x_{ij}^{T}\beta})\\
    =&-\frac{1}{2}\sum_{i=1}^{n}\sum_{j=1}^{m_{i}}w_{ij}^{t}[u_{ij}^{t}-\beta_{0}-x_{ij}^{T}\beta]^{2}+C+R_{2}(\beta_{0},\beta\;|\;\beta_{0}^{t},\beta^{t})\\
    \equiv & Q_{Q}(\beta_{0},\beta\;|\;\beta_{0}^{t},\beta^{t})+ C+R_{2}(\beta_{0},\beta\;|\;\beta_{0}^{t},\beta^{t})
\end{aligned}\]
where $C$ is a constant which is independent of $\beta_{0}$ and
$\beta$; $R_{2}(\beta_{0},\beta\;|\;\beta_{0}^{t},\beta^{t})$ is the
remainder term;
\[
    u_{ij}^{t} =\beta_{0}^{t}+x_{ij}^{T}\beta^{t}+
        \frac{z_{i}\gamma_{ij}^{t}-p_{ij}^{t}}{p_{ij}^{t}q_{ij}^{t}},
    \quad w_{ij}^{t} =p_{ij}^{t}q_{ij}^{t},
\]
$p_{ij}^{t}
=\left[1+e^{-(\beta^t_{0}+x_{ij}^{T}\beta^t)}\right]^{-1}$, and
$q_{ij}^t=1-p_{ij}^t$. Using this quadratic approximation
$Q_{Q}(\beta_{0},\beta\;|\;\beta_{0}^{t},\beta^{t})$, computing time
can be boosted up to $20$ times faster than the program without
using approximation. Hereafter, we work on
$Q_{Q}(\beta_{0},\beta\;|\;\beta_{0}^{t},\beta^{t})$ rather than
$Q(\beta_{0},\beta\;|\;\beta_{0}^{t},\beta^{t})$.

In the M-step, we have to solve the following maximization problem,
\[
    \max_{\beta_{0},\beta}Q_{Q}(\beta_{0},\beta\;|\;\beta_{0}^{t},\beta^{t}).
\]
Since $Q_{Q}(\beta_{0},\beta\;|\;\beta_{0}^{t},\beta^{t})$ is a
quadratic function of $\beta_0$ and $\beta$, the maximization
problem is equivalent to finding the root of
\begin{equation} \label{eq:NE}
    \frac{\partial}{\partial\beta_{k}}Q_{Q}(\beta_{0},\beta\;|\;\beta_{0}^{t},\beta^{t})
    =\sum_{i=1}^{n}\sum_{j=1}^{m_{i}}w_{ij}^{t}x_{ij,k}(u_{ij}^{t}-\beta_{0}-x_{ij}^{T}\beta)=0
\end{equation}
{for all $k = 1, \dots, p$, where $x_{ij,0}=1$ for all $i,j$,
and $\beta_{k}$ is the $k$th element of $\beta$. Here we adopted
coordinate decent algorithm (updating one coordinate at a time)
proposed in \cite{F10}. Since (\ref{eq:NE}) is a linear in terms of
$\beta_k$'s, the updating formula for $\beta_{k}$ is straight
forward.  At step $t+1$, let
$S_{0}=\sum_{i=1}^{n}\sum_{j=1}^{m_{i}}w_{ij}^{t}(u_{ij}^{t}-x_{ij}^{T}\beta^{t})$
and
$$S_{k}=\sum_{i=1}^{n}\sum_{j=1}^{m_{i}}w_{ij}^{t}x_{ij,k}(u_{ij}^{t}-\beta_{0}^{t}-x_{ij}^{T}\beta_{(k)}^{t}),$$
where $k=1,\dots,p$, and $\beta_{(k)}^{t}$ is $\beta^{t}$ with its
$k$th element replaced by 0. The updating formula is
\[
    \beta_{k}^{t+1}=\frac{S_{k}}{\sum_{i=1}^{n}\sum_{j=1}^{m_{i}}
    w_{ij}^{t}(x_{ij,k})^{2}} \quad (k=0,\dots,p).
\]

\section{Penalized Multiple-Instance Logistic Regression}

In the manufacturing process, one important issue is to
identify the active factors within the process, especially for large
$p$. Traditionally the stepwise procedure is used to search the
active covariates and after identifying these important covariates,
the coefficients of these covariates are estimated based on the
current model. Here we want to integrate the maximum likelihood
coefficient estimation and the variable selection into one single
procedure. Thus the idea is to shrike the small coefficient values
to be zeros. Therefore LASSO type method \cite{T96} is
adopted.

In order to perform estimation and variable selection at the same
time, we include LASSO penalty into our model to shrink the
unimportant coefficients to zero. In this work, the intercept
$\beta_{0}$ is always kept in the model.  The resulting optimization
problem is therefore
\[
    \min_{\beta_{0},\beta}\left\{-Q_{Q}(\beta_{0},\beta\;|\;\beta_{0}^{t},\beta^{t})
        +\lambda\sum_{k=1}^{p}|\beta_{k}|\right\}.
\]
Shooting algorithm \cite{F98} is efficient to solve this optimization problem.
The resulting updating formula is
\[\begin{aligned}
    &\beta_{0}^{t+1}= \frac{S_{0}}{\sum_{i=1}^{n}\sum_{j=1}^{m_{i}}w_{ij}^{t}}
    \quad \mbox{and}\\
    &\beta_{k}^{t+1}=  \left\{ \begin{array}{lll}
        (S_{k}-\lambda)/\sum_{i=1}^{n}\sum_{j=1}^{m_{i}}w_{ij}^{t}(x_{ij,k})^{2}
        & \mbox{if} & S_{k}>\lambda\\
        (S_{k}+\lambda)/\sum_{i=1}^{n}\sum_{j=1}^{m_{i}}w_{ij}^{t}(x_{ij,k})^{2}
        & \mbox{if} & S_{k}<-\lambda\\
        0 & \mbox{if} & |S_{k}|\leq\lambda
    \end{array} \right.
\end{aligned}\]
for $k=1,\dots, p$.

To choose the optimal tuning parameter $\lambda$, we first determine the upper bound of $\lambda$, say $\lambda_{max}$ which enforces $\beta$ to be $0$. We notice that $\gamma_{ij}^t > p_{ij}^t$ and
\begin{equation} \label{eq:main}
    \left| z_i \gamma_{ij}^t-p_{ij}^t \right| \leq \left\{ \begin{array}{llll}
    p_{ij}^t \left( \frac{[1-p_{ij}^t]^{m_i}}{1-[1-p_{ij}^t]^{m_i}}\right)
    & \leq & m_i^{-z_i} & \mbox{ if } z_i=1 \\
    p^t_{ij} & \leq & 1 & \mbox{ if } z_i=0 \end{array} \right..
\end{equation}
So when $\beta^t=0$, we have, for any $k=1,\dots,p$,
\[\begin{aligned}
    &\sum_{ij}w_{ij}^{t}x_{ij,k}(u_{ij}^{t}-\beta_{0}^{t}-x_{ij}^{T}\beta_{(k)}^{t})
    = \sum_{ij}x_{ij,k} (z_i\gamma_{ij}^t - p_{ij}^t)\\
    \leq & \left[ \sum_{ij}x^2_{ij,k}\right]^{1/2}\left[ \sum_{ij}m_i^{-2z_i}\right]^{1/2}\\
    = & \left[\sum_{i=1}^n (m_i - 1)\right]^{1/2} \left[\sum_{i=1}^n m_i^{1-2z_i}\right]^{1/2} \equiv \lambda_{max}
\end{aligned}\]
where the first inequality is due to Cauchy-Schwarz inequality and (\ref{eq:main}), and the equality right next to the inequality is due to that $x_k$'s are normalized prior to data analysis.

Several technical details are crucial to end up with an automatic parameter tuning. We follow the suggestion of \cite{F10} to adjust our computer codes. First, we choose a sequence of  $\lambda$, ranging from $\lambda_{min}=\epsilon\lambda_{max}$ to $\lambda_{max}$ in a descending order, say $\lambda_1 < \lambda_2 < \cdots < \lambda_K$. Set $\epsilon=0.001$ and the length of the sequence $K=20$. The optimal $\lambda$ is chosen among these $K$ values. Second, when $p_{ij}^tq_{ij}^t$ is too small, the value of $u_{ij}^t$ stored in computer may deviate from the true greatly. In this sequel, when $p_{ij}^t > 1- 10^{-5}$ we set $p_{ij}^t = 1$ and $w_{ij}^t=p_{ij}^tq_{ij}^t=10^{-5}$ and when $p_{ij}^t < 10^{-5}$ we set $p_{ij}^t = 0$ and $w_{ij}^t=p_{ij}^tq_{ij}^t=10^{-5}$. Finally, we choose the best tuning parameter by $\kappa$-fold cross validation. The procedure for choosing tuning parameter applied in this note is
\begin{enumerate}
    \item[] FOR $i$ in the sequence of $\lambda$'s
        \begin{enumerate}
            \item[] Randomly split the data into $\kappa$ subsets used for $\kappa$-fold cross-validation
            \item[] FOR $j=1$ to $K$
                \begin{enumerate}
                    \item Estimate the parameters using $\lambda=\lambda_i$ and the whole data except for the $j$th subset
                    \item Compute deviance = $-2\log likelihood$ using the estimated parameters and the $j$th subset
                \end{enumerate}
            \item[] END FOR
            \item[] Compute the mean and standard error of the 10 deviances
        \end{enumerate}
    \item[] END FOR
    \item[] Choose the optimal tuning parameter as the $\lambda$ with the smallest mean deviance
\end{enumerate}
For demonstration, we set $n=100$, $m_i=3$, $p=100$ with only 5 out of them are active and $\kappa=10$. The results are shown in Figures 1 and 2. The optimal $\lambda_{opt}$ selected via deviance is 2.31. 

\begin{figure}[!t]
    \centering
    \includegraphics[width=2.5in]{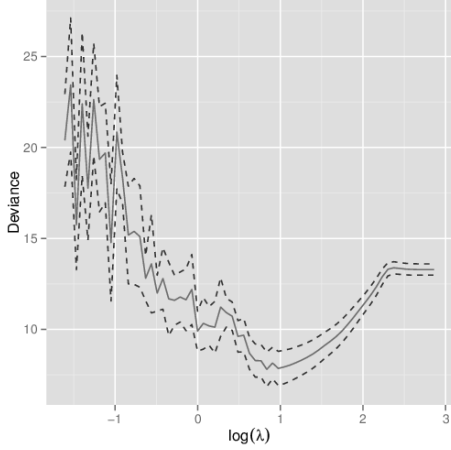}
   \caption{Ten-fold cross-validation on the simulated data sets. The red line is the mean deviance and the blue lines are bounds for deviances within one standard error.}
\end{figure}    
    
\begin{figure}[!t]
	\centering
	\includegraphics[width=2.5in]{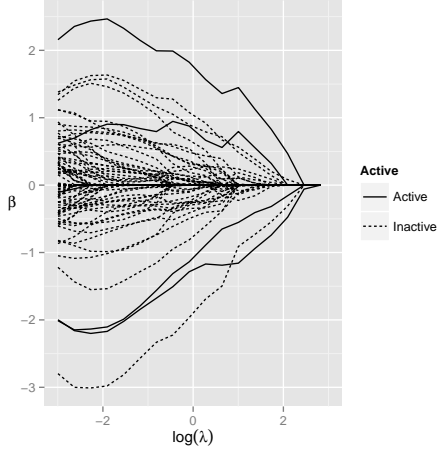}
    \caption{The change of the estimated parameters with respect to the tuning parameter. The red lines stands for active covariates and the blue lines stands for inactive covariates.}
\end{figure}


\section{Simulation Studies}
\subsection{Naive vs MILR}
To demonstrate the powerfulness of the proposed model, we consider a simulation with data generating process as addressed in Section 4. We generated 100 datasets with $n=100$, $m=3$,  $\beta_0=-2$, and $\beta=(1,-1,0)$. That is, we only generate 3 covariates and the third covariate is inactive to the response. Simulation results are summarized in Table \ref{tb:simu1} which shows that using the MILR results in unbiased estimations and more powerful (Wald) tests than using the naive method. As shown in Table \ref{tb:simu1}, the MLEs of MILR are empirically unbiased and more powerful in contrast to the naive method. Especially, the naive estimates of regression coefficients were severely attenuated which may result in relatively high prediction errors. This says that if the goal of data analysis is to identify important covariates then the Naive and the MILR approach may not yield drastically different results. However, if the goal is to predict whether change of one particular covariate can reduce the chance of being defect, then the naive approach may mislead the result.  

\begin{table}[ht]
\centering
\caption{(Average Estimate, Standard Error, Power) of Regression Coefficient Estimation / Testing
 \label{tb:simu1}}
\begin{tabular}{ccccc}
  \hline
  Method & $\beta_0=-2$ & $\beta_1=1$ & $\beta_2=-1$ & $\beta_3=0$ \\
  \hline
  Naive & (-0.19, 0.02, & (0.34, 0.01, & (-0.33, 0.01, & (0.01, 0.01, \\
  	& 0.39) & 0.80) & 0.77) & 0.03)\\
  MILR & (-2.29, 0.05, & (1.28, 0.06, & (-1.02, 0.04, & (0.04, 0.03, \\
  	& 0.93) & 0.86) & 0.87) & 0.06)\\
   \hline
\end{tabular}
\end{table}

\subsection{MILR-LASSO}
In this example, we demonstrate the performance of the proposed method for large $p$ small $n$ cases. The data generating process is designed as follows. We generated $B=50$ independent datasets. Each dataset consists of $n$ subjects and each subject consists of $m$ components. The response of the $j$th component nested in the $i$th subject, $Y_{ij}$, follows $Ber(p(\beta_0 + x_{ij}^T\beta))$ where $x_{ij} \in \Re^p$ is a $p$-dimensional vector randomly and independently sampled from the standard normal distribution. The response of the $i$th subject is defined as $Z_i=I(\sum_{j=1}^m Y_{ij}>0)$. We simulated data with all possible configurations of factor $n=100$ and $m=3$. The number of predictors $p$ is $100$ excluding the intercept. For each data set, we randomly assigned $-2,-1,1,2,0.5,$ and 95 multiples of 0 to the regression coefficient of predictors. Last, $\lambda_{opt}$ is the value which minimizes the deviance using 10-fold cross-validation. Following three variable selection schemes are considered:
\begin{enumerate}
\item[(A)] MILR model with LASSO penalty;
\item[(B)] MILR model with forward selection using Wald test with $\alpha=0.05$; and
\item[(C)] Naive model with forward selection using Wald test with $\alpha=0.05$.
\end{enumerate}

\begin{table}[ht]
\centering
\caption{Variable Selection Results (n=100, m=3)\label{tb:simu2}}
\begin{tabular}{c|cccc}
  \hline
  Model &  True Positive & False Positive & True Negative & False Negative \\
  \hline
  (A) & 0.78 & 0.15 & 0.85 & 0.22\\
  (B) & 0.72 & 0.06 & 0.94 & 0.28\\
  (C) & 0.58 & 0.07 & 0.93 & 0.42\\
  \hline
\end{tabular}\\
\end{table}

\subsection{Compare with other methods}
To compare our MILR-LASSO with other MILR methods (MILR-s(3) from \cite{RC05} and MILR-s(0) from \cite{XF04}), we designed three different simulation schemes. The first scheme consider fixed $m=5$; the second consider various $m_i$ with mean 5; and the third consider various $m_i$ with mean 65. These schemes mimicked the real datasets MUSK1 and MUSK2 which will be introduced in Section V. Some summary statistics about these datasets are listed in Table \ref{tb:info}. The regression coefficients used to generate these simulated data sets are the estimated coefficients of the MUSK data sets using MILR-LASSO model. Hence, most of the coefficients are zeros (about only $5\%$ of the coefficients are non-zero). Besides using 10-fold cross-validation to select the optimal $\lambda$, we also choose BIC to obtain optimal LASSO model which is more efficient for it only need a single fit \cite{ZHT07}. The following are three different simulation schemes:
\begin{itemize}
\item[(D)] $n=100$, $p=166$, $m_i=5$ for all $i=1,...,n$
\item[(E)] $n=100$, $p=166$, $m_i \sim Poisson(4) + 1$ (similar to MUSK1 dataset shown in Case Studies )
\item[(F)] $n=100$, $p=166$, $m_i \sim Poisson(64) + 1$ (similar to MUSK2 dataset shown in Case Studies)
\end{itemize}
We set $m_i \sim Poisson(4) + 1$ instead of $m_i \sim Poisson(5)$ to avoid the case of $m_i=0$. Then we used 10-fold stratified cross-validation to test these algorithms. To generate the subject-level prediction from the estimated coefficients $(\hat{\beta_0}, \hat{\beta})$ of these algorithms, we use $0.5$ as a threshold. Thus,
\[
    \hat{Z_i} = I(1-\prod_{j=1}^{m_i}(1-\hat{p}_{ij}) \geq 0.5),
\]
where $\hat{p}_{ij} = p(\hat{\beta_0}+x^T_{ij}\hat{\beta})$. To evaluate these three algorithms, two summary statistics are reported:  accuracy (ACC), and the area under the ROC curve (AUC). Also, each algorithm is repeated $B=100$ times.

\begin{table}[ht]
\centering
\caption{Predicted results of different methods on simulated data sets \label{tb:simu3}}
\begin{tabular}{llcc}
  \hline
  Scheme & Method &  ACC & AUC \\
  \hline
(D) & MILR-LASSO(BIC) & 0.61(0.008) & 0.62(0.012) \\
& MILR-LASSO(10-fold CV) & 0.64(0.008) & 0.61(0.014) \\
& MILR-s(3) & 0.58(0.004) & 0.57(0.009) \\
& MILR-s(0) & 0.58(0.005) & 0.57(0.009) \\
  \hline
(E) & MILR-LASSO(BIC) & 0.70(0.007) & 0.75(0.009) \\
& MILR-LASSO(10-fold CV) & 0.70(0.007) & 0.75(0.009) \\
& MILR-s(3) & 0.58(0.004) & 0.59(0.008) \\
& MILR-s(0) & 0.58(0.004) & 0.58(0.009) \\
  \hline
(F) & MILR-LASSO(BIC) & 0.82(0.003)& 0.53(0.006)\\
& MILR-LASSO(10-fold CV) & 0.82(0.003)& 0.53(0.007)\\
& MILR-s(3)  & 0.18(0.003)& 0.46(0.009)\\
& MILR-s(0)  & 0.18(0.003)& 0.45(0.011)\\
  \hline
\end{tabular}\\
\end{table}

From Table \ref{tb:simu3}, it is obvious that MILR-LASSO outperforms the other two methods, since over $95\%$ of the predictors are nuisance variables. Furthermore, both BIC and 10-fold CV provide similar prediction results. Thus, for the sake of efficiency, we prefer using BIC to find the optimal model.

\section{Case Studies}
The MUSK data sets are the most widely used benchmark data sets when comparing different MI learning methods. The MUSK datasets consist of conformations. Each conformation is represented by 166 features. In MUSK1, the average conformation in one bag is 5; whereas, in MUSK2, the average conformation in one bag is 65. More descriptive statistics about these datasets are shown in Table \ref{tb:info}. For more detailed descriptions, please refer to \cite{D97}.

\begin{table}[ht]
\centering
\caption{Information about these two data sets\label{tb:info}}
\begin{tabular}{c|cccccc}
  \hline
  Data set &  $p$ & $\bar{m}^{\dagger}$ & $n$ & Sample Size & Prop. of Positive Subj\\
  \hline
 MUSK1 & 166 & 5.17 & 92 & 476 & 51.08\%\\
 MUSK2 & 166 & 64.69 & 102 & 6598 & 38.24\%\\
  \hline
\end{tabular}\\
$\dagger$: the average bag size.
\end{table}

To apply MILR-LASSO on these data sets, first we choose the tuning parameter ($\lambda$) via 10 fold cross-validation. Then, use the selected $\lambda$ to fit the data sets. To avoid the likelihood from being only locally optimal, we replicate each algorithm 10 times and average their performance. The right halves of the Tables \ref{tb:case1fit} and Table \ref{tb:case2fit} are the predicted results (10-fold cross-validation) of three different algorithms and the left halves of Table \ref{tb:case1fit} and Table \ref{tb:case2fit} are the fitted results.

\begin{table}[]
\centering
\caption{Fitted and predicted results of different methods on MUSK1 data set\label{tb:case1fit}}
\begin{tabular}{lcccc}
  \hline
  & \multicolumn{2}{c}{Fitted} & \multicolumn{2}{c}{Predicted}\\
Method & ACC & AUC & ACC & AUC\\
  \hline
MILR-LASSO & 1.00 & 1.00 & 0.79 & 0.83 \\
MILR-s(3) &  0.85 & 0.96 & 0.72 & 0.76 \\
MILR-s(0) &  0.87 & 0.93 & 0.74 & 0.79 \\
  \hline
\end{tabular}\\
\end{table}

\begin{table}[]
\centering
\caption{Fitted and predicted results of different methods on MUSK2 data set\label{tb:case2fit}}
\begin{tabular}{c|cccccc}
  \hline
   & \multicolumn{2}{c}{Fitted} & \multicolumn{2}{c}{Predicted}\\
  Method &  ACC & AUC & ACC & AUC \\
  \hline
 MILR-LASSO & 0.87 & 0.96 & 0.69 & 0.76\\
MILR-s(3) & 0.99 & 1.00 & 0.74 & 0.83\\
MILR-s(0) & 0.95 & 1.00 & 0.79 & 0.85\\
  \hline
\end{tabular}\\
\end{table}

From Tables \ref{tb:case1fit} and \ref{tb:case2fit}, we can see that no algorithm is consistently better than the others. However, MILR-LASSO has the strength to select important features when estimating the coefficients.

\section{Conclusion}
In this work, the multiple instance learning is treated as a classical missing value problem and solved by EM algorithm. In addition, the lasso approach is applied to identify important covariates. This treatment allows us to figure out influential covariates, to predict defect rate, and, most importantly, to direct ways to potentially reduce the defect rate by adjusting covariates. The limitations of the proposed method are as follows. First, we ignore the potential dependency among observations within a subject. Random effects can be incorporated into the proposed logistic regression to represent the dependency. Second, in a preliminary simulation study, not shown in this paper, we discovered that the maximum likelihood estimator is biased under the model (F). Bias reduction methods such as  \cite{Q56} and \cite{F93} will be applied in our future work.

\end{document}